%% file: main.tex
\documentclass[10pt,twocolumn,letterpaper]{article}

\usepackage[pagenumbers]{wacv} 

\usepackage{adjustbox}
\usepackage{algorithm}
\usepackage{algpseudocode}
\usepackage{wacv}
\usepackage{times}
\usepackage{epsfig}
\usepackage{graphicx}
\usepackage{amssymb}
\usepackage{booktabs}
\usepackage{amsmath,multirow,amsfonts,booktabs,xcolor}
\usepackage{array}
\usepackage{textcomp}
\usepackage{stfloats}
\usepackage{url}
\usepackage{verbatim}
\usepackage{enumitem}
\usepackage{colortbl}
\floatstyle{plain}
\restylefloat{figure}

\definecolor{darkgreen}{RGB}{0,200,0}
\usepackage[pagebackref,breaklinks,colorlinks]{hyperref}

\DeclareMathOperator*{\argmin}{arg\,min}

\usepackage[capitalize]{cleveref}
\crefname{section}{Sec.}{Secs.}
\Crefname{section}{Section}{Sections}
\Crefname{table}{Table}{Tables}
\crefname{table}{Tab.}{Tabs.}

\def\wacvPaperID{217} 
\def\confName{WACV}
\def\confYear{2024}

\begin{document}

\title{Back to Optimization: Diffusion-based Zero-Shot 3D Human Pose Estimation}

\author{
Zhongyu Jiang$^{1*}$ \quad 
Zhuoran Zhou$^{1}$\thanks{~Equal contribution.} \quad 
Lei Li$^2$ \quad 
Wenhao Chai$^1$ \\
Cheng-Yen Yang$^1$ \quad 
Jenq-Neng Hwang$^1$ \\
[2mm]
University of Washington$^1$ \quad 
University of Copenhagen$^2$\\
[2mm]
\tt\small \{zyjiang, zhouz47, wchai, cycyang, hwang\}@uw.edu, lilei@di.ku.dk
}
\maketitle

\begin{abstract}
Learning-based methods have dominated the 3D human pose estimation (HPE) tasks with significantly better performance in most benchmarks than traditional optimization-based methods. Nonetheless, 3D HPE in the wild is still the biggest challenge for learning-based models, whether with 2D-3D lifting, image-to-3D, or diffusion-based methods, since the trained networks implicitly learn camera intrinsic parameters and domain-based 3D human pose distributions and estimate poses by statistical average. On the other hand, the optimization-based methods estimate results case-by-case, which can predict more diverse and sophisticated human poses in the wild.
By combining the advantages of optimization-based and learning-based methods, we propose the \textbf{Ze}ro-shot \textbf{D}iffusion-based \textbf{O}ptimization (\textbf{ZeDO}) pipeline for 3D HPE to solve the problem of cross-domain and in-the-wild 3D HPE. Our multi-hypothesis \textit{\textbf{ZeDO}} achieves state-of-the-art (SOTA) performance on Human3.6M, with minMPJPE $51.4$mm, without training with any 2D-3D or image-3D pairs. Moreover, our single-hypothesis \textit{\textbf{ZeDO}} achieves SOTA performance on 3DPW dataset with PA-MPJPE $40.3$mm on cross-dataset evaluation, which even outperforms learning-based methods trained on 3DPW.
\end{abstract}

\section{Introduction}

As people become increasingly interested in Virtual Reality (VR), Augmented Reality (AR), Human-Computer Interaction and Sports Analysis, 3D Human Pose Estimation (HPE) becomes a crucial component for these applications.
Compared with multi-view 3D HPE, monocular-based methods are easier to set up and have lower costs, which are more suitable for VR, AR, and mobile devices. Weng \etal\cite{weng2022humannerf}, and Peng \etal\cite{peng2021neuralbody} utilize 3D human poses with Neural Radiance Fields (NeRF) for 3D Human Reconstruction. Meanwhile, Bridgeman \etal\cite{bridgeman2019multisoccer} propose a 3D HPE and tracking pipeline for soccer analysis, and Jiang \etal\cite{jiang2022golfpose} take advantage of 2D and 3D HPE to track the motion of golf players.

With the availability of more benchmark datasets, deep learning-based 3D HPE methods have been shown to outperform all traditional methods and dominate the areas. Combining 2D HPE with SMPL \cite{SMPL} model, Bogo \etal propose SMPLify\cite{bogo2016smplify} as an optimization-based 3D HPE pipeline. 2D-3D lifting \cite{videopose3d, SemGCN, ci2020locally, camerapose, poseformer} and diffusion-based 3D HPE\cite{ci2023gfpose, gong2023diffpose} networks leverage 3D human poses from the single-frame or multi-frame 2D poses. On the other hand, Image-to-3D networks \cite{SPIN, kocabas2020vibe, BEV, CLIFF} estimate 3D human poses directly from images without intermediate 2D human poses. However, as mentioned in \cite{poseaug, gholami2022adaptpose}, learning-based 3D HPE methods suffer from performance degradation with cross-domain or in-the-wild scenarios. During training, these methods implicitly learn camera intrinsic parameters, domain-based 3D human pose distributions, or image features in a certain domain. Although optimization-based methods can mitigate the impact of domain gaps by estimating 3D poses case-by-case, their performances are not comparable to learning-based methods at this moment. 

\begin{figure}[t]
    \centering
    \includegraphics[width=\linewidth]{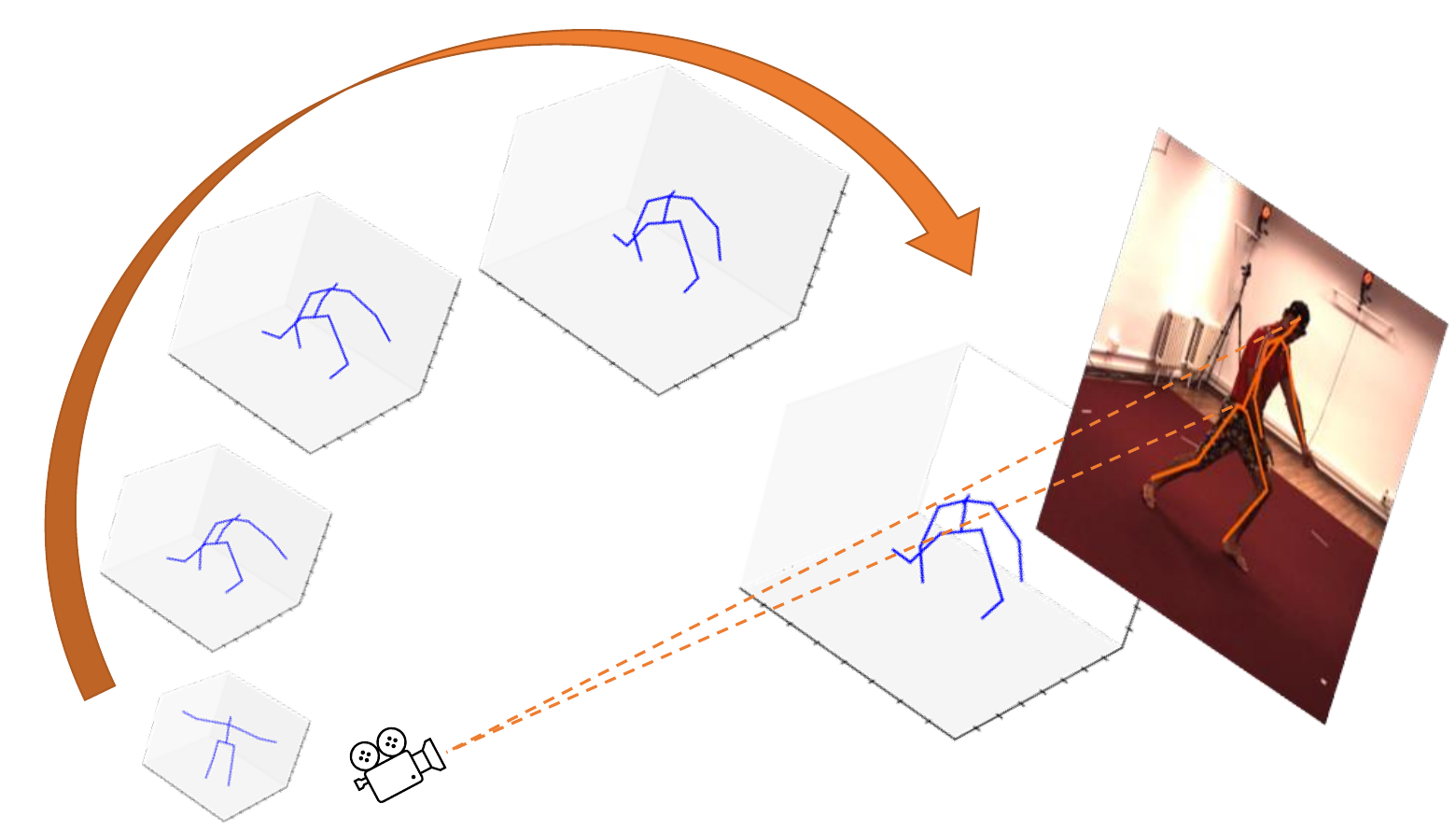}
    \caption{ZeDO iteratively estimates 3D poses by minimizing the re-projection error via a diffusion-based method.}
    \label{fig:enter-label}
\end{figure}

To address this problem, Zhan \etal\cite{zhan2022ray3d} decouple the camera intrinsic parameters from 2D-3D lifting networks learning by converting 2D keypoints to 2D rays. Gong \etal\cite{poseaug} and Gholami \etal\cite{gholami2022adaptpose} generate various 3D poses to bridge domain gaps. Furthermore, Chai \etal\cite{chai2023global} propose a data augmentation pipeline to minimize the 3D human pose spatial distribution gap. However, the above methods still cannot outperform the learning-based 3D HPE. To decouple camera intrinsic and bridge 3D pose domain gaps simultaneously, we propose the Zero-shot Diffusion-based Optimization(ZeDO) pipeline for 3D human pose estimation, which combines a simple yet effective optimization pipeline with a pre-trained diffusion-based 3D human pose generation model. 

Different from traditional optimization-based methods\cite{bogo2016smplify, thomas2018joint, renshu2022unsupervised, muller2021SMPLify-XMC}, which include various kinematic constraints, the diffusion model denoises the output from the optimization pipeline iteratively to ensure optimized poses following human body constraints. Meanwhile, poses are optimized by minimizing 2D keypoint re-projection errors with a simple yet effective optimization pipeline. Thus, ZeDO is able to estimate 3D human poses without training on any 2D-3D or image-3D pairs. Our contributions are as follows:
\begin{itemize}
    \item The proposed ZeDO pipeline is a Zero-Shot 3D Human Pose Estimation pipeline, which leverages a pre-trained diffusion-based 3D human pose generation model to optimize target 3D poses in the loop during the inference time.
    \item Compared with other generation and denoising tasks, we take the diffusion model as an optimization tool by combining a simple 3D HPE optimization pipeline with a pre-trained diffusion-based 3D human pose generation model.
    \item ZeDO achieves state-of-the-art zero-shot 3D HPE performance on Human3.6M, 3DHP, and 3DPW datasets, even on cross-dataset evaluation.
\end{itemize}

In Sec~\ref{sec:related}, we will discuss related works in 3D HPE, including optimization-based and learning-based methods. In Sec~\ref{sec:method}, details of our backbone architecture and optimization pipeline are addressed. Experimental results are presented in Sec~\ref{sec:exp}, and the ablation studies will be discussed in Sec~\ref{sec:discuss}. At last, there are conclusions in Sec~\ref{sec:conclusion}.

\section{Related Works} 
\label{sec:related}

\subsection{Optimization-based Human Pose Estimation}
Optimization-based methods, which estimate 3D poses frame-by-frame and case-by-case, are not handicapped by domain gaps or varying camera intrinsic parameters, but their performance so far is much worse than learning-based methods. Utilizing the SMPL\cite{SMPL} model, SMPLify\cite{bogo2016smplify} is capable of optimizing the 3D human poses by minimizing the 2D keypoint re-projection error and satisfying lots of kinematic constraints. Furthermore, M{\"u}ller \etal\cite{muller2021SMPLify-XMC} propose SMPLify-XMC as an improved version of SMPLify with more constraints about the human body and more inputs including height and age. Zheng \etal\cite{thomas2018joint} propose an optimization-based hierarchical 3D human pose estimation pipeline that can estimate both 3D human pose and locations at the same time. 
Recently, Song \etal\cite{song2020learnedgradient} and Choutas \etal\cite{choutas2022FitMorphable} focus on train an optimizer to fit the SMPL model to estimated 2D human poses.

\subsection{2D-3D Lifting Networks}
2D-3D lifting networks employ either a single frame or a sequence of normalized 2D keypoints as input to generate corresponding 3D keypoints\cite{simplebaseline, videopose3d, camerapose, SemGCN}. Pavllo \etal\cite{videopose3d} leverage dilated temporal convolutions with semi-supervised ways to improve 3D pose estimation in videos. Yang \etal\cite{camerapose} enhance 3D human pose estimation by leveraging in-the-wild 2D annotations and a novel refinement network module in a weakly-supervised framework. Li \etal\cite{evoskeleton} propose a scalable data augmentation technique that synthesizes unseen 3D human skeletons for training 2D-to-3D networks, effectively reducing dataset bias and improving model generalization to rare poses. These methods, despite the need of two-stage processing to obtain the 2D keypoints in advance, have demonstrated superior performance on several benchmark datasets and are highly efficient, especially when adapted for temporal considerations.

\subsection{Image-to-3D Networks}

End-to-end 3D HPE methods directly transform image data into 3D pose representations, such as those put forth by Guler \etal\cite{guler2019holopose}, Tung \etal\cite{tung2017self}, Tan \etal\cite{tan2017indirect}, and various other research teams including those behind SPIN\cite{SPIN}, ROMP\cite{romp}, BEV\cite{BEV}, and CLIFF\cite{CLIFF}, who successfully utilize scale and variable height information, effectively resolving issues of depth/height ambiguity. For instance, the methodology introduced by Sun \etal\cite{romp} is a one-stage process that allows for real-time, monocular 3D mesh recovery of multiple individuals. Further contributing to this field, Sun \etal\cite{BEV} develop a single-shot method capable of simultaneously regressing the pose, shape, and relative depth of multiple people within a single image, utilizing the Bird's-Eye-View representation for depth reasoning while accommodating variable heights. Within the area of 3D pose estimation from single images, these one-stage techniques consistently demonstrate robust performance, despite their comparatively streamlined architectural designs. 

\subsection{Diffusion Models in Human Pose Estimation.} 
Nowadays, Diffusion Probabilistic Model\cite{pmlr-v37-sohl-dickstein15} and its descendants\cite{ho2020denoising, rombach2022high, song2020denoising} have shown their outstanding capabilities of prior distribution learning in multiple areas, such as image generation\cite{ramesh2021zero} and editing\cite{cao2023difffashion, cao2023image}, 3D-reconstruction\cite{poole2022dreamfusion}, image inpainting\cite{lugmayr2022repaint}, and human motion generation\cite{tevet2022human}. Intuitively, diffusion models are able to tackle depth/scale ambiguity and one-to-many mappings in 3D HPE tasks. Ci \etal\cite{ci2023gfpose} introduce a novel score-based generative framework to model plausible 3D human poses with a hierarchical condition masking strategy. Meanwhile, Gong \etal\cite{gong2023diffpose} propose a diffusion-based architecture including the initialization of 3D pose distribution, a GMM-based forward diffusion process, and a conditional reverse diffusion process.

Although learning-based methods achieve significant success in 3D HPE tasks, they are highly constrained by the training datasets. In other words, their performance would abruptly drop when tested on other datasets due to domain gaps. To resolve this issue, we integrate the generality of optimization methods into a robust pre-trained diffusion backbone and propose a Zero-shot Diffusion-based Optimization pipeline (ZeDO) for 3D HPE.

\section{Method}
\label{sec:method}

\begin{figure*}[!ht]
\begin{center}
    \includegraphics[width=0.9\linewidth]{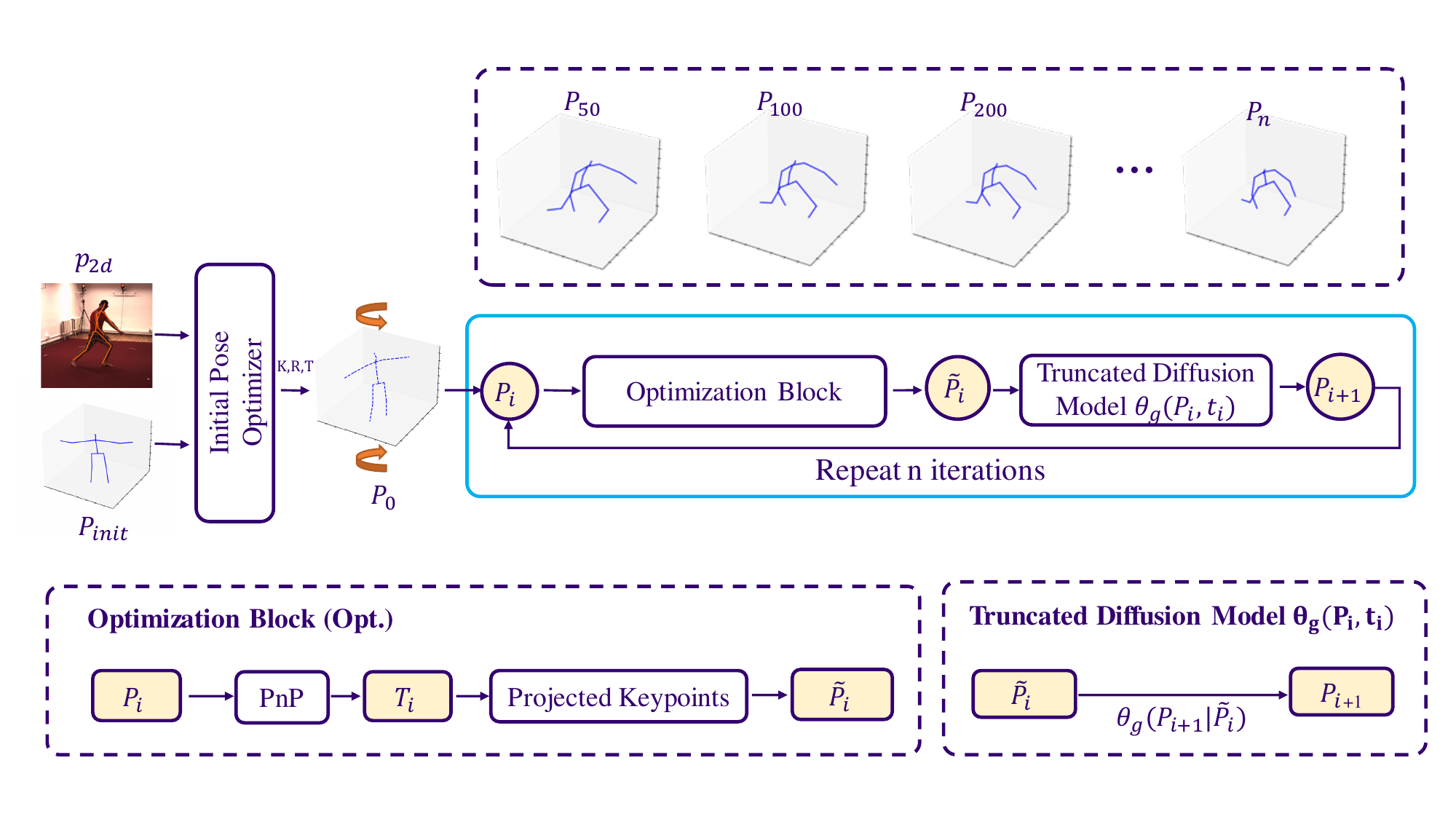}
\end{center}
   \caption{The pipeline of ZeDO, which takes an initial 3D pose, called  a hypothesis, as input and estimates the pose by minimizing re-projection error with the target 2D pose. After 1000 iterations, ZeDO is able to generate the optimized 3D human poses.}
\label{fig:pipeline}
\end{figure*}

As shown in Fig~\ref{fig:pipeline}, ZeDO includes an initial pose optimizer for rotating initial poses and an optimizer in the loop for iteratively optimizing 3D poses.

Firstly, a randomly selected initial 3D pose (a hypothesis), $P_{init} \in R^{J\times3}$, is rotated to an optimal pose, $P_{0}$, by minimizing the re-projection error with detected or ground truth 2D keypoints  $p_{2d} \in R^{J\times2}$. Here, $J$ stands for the number of keypoints. Then, in the $i$th optimization step, $P_{i}$ is optimized by the optimizer in the loop, and the pre-trained diffusion model is used to denoise it to $P_{i+1}$ as input of the next iteration. After $n$ iterations, $P_{n}$ will be the estimated 3D human pose.

Different from other diffusion-based pose estimation methods\cite{ci2023gfpose, gong2023diffpose}, our diffusion model $\theta_g$ is a pose generation model, which is only trained with 3D human poses, and during inference, our diffusion model only takes the optimized pose $\widetilde{P}_i$ and timestamp $t(i)$ as input without any additional pose condition information including 2D poses.

\subsection{Pre-trained 3D Human Pose Generation Model}
We apply the Score Matching~\cite{song2020score} on the pre-trained backbone for our 3D human pose generation diffusion model, which rectifies the noisy poses generated after projection to get reasonable 3D poses. During pre-training, the model takes relative-to-pelvis 3D poses $x \in R^{J\times3}$ as inputs and tries to recover them from recurrent Gaussian noise. In this case, we expect that the diffusion model learns the distribution of real 3D poses and reconstructs $\tilde{x} \in R^{J\times3}$ to minimize the difference from the inputs.  The perturbation strategy used in our Score Matching diffusion model expresses $p(x(t)|x(0))$ in the closed form as:
\begin{equation}
 N\Big(x(t);x(0)e^{-\frac{1}{2} \int_{0}^t \beta(s) ds},\big[1-e^{-\int_{0}^{t}\beta(s)ds}\big]^2 I\Big) \tag{1}.
\end{equation}

Besides, built upon the learning strategy of the noise conditional score network (NCSN) ~\cite{song2019generative}, we formulate our loss function as follows by choosing $\lambda(t) = \sigma(t)^2$:
\begin{align}
    L &= E_{U(t;0,1)} \Big[\lambda(t)||s_{\theta}(x(t),t)+\frac{x(t)-\mu}{\sigma^{2}}||^{2}_{2}\Big] \tag{2} \\
&= E_{U(t;0,1)} \Big[||\sigma(t)s_{\theta}(x(t),t)+z||^{2}_{2}\Big] \tag{3},
\end{align}

\noindent in which $z$ stands for random noise vector $z \sim N\left(0,1\right)$ and $s_{\theta}$ is the pre-trained score matching network. $\sigma$ represents the variance mentioned in Eq.$(1)$ as $[1-e^{-\int_{0}^{t}\beta(s)ds}]$ and the timestamp or denoising time variable $t$ is uniformly sampled 1000 times from $\left(0, 1\right]$. The 3D pose generation model is never trained with any 2D-3D or image-3D pairs. We report the results based on the Score Matching based model, similar to GFPose\cite{ci2023gfpose}. More results with different backbones, including DDIM\cite{song2020denoising} and DDPM\cite{2020DDPM}, are shown in Table~\ref{tab:diffbb}.

\begin{figure}[t]
    \centering
    \includegraphics[width=\linewidth]{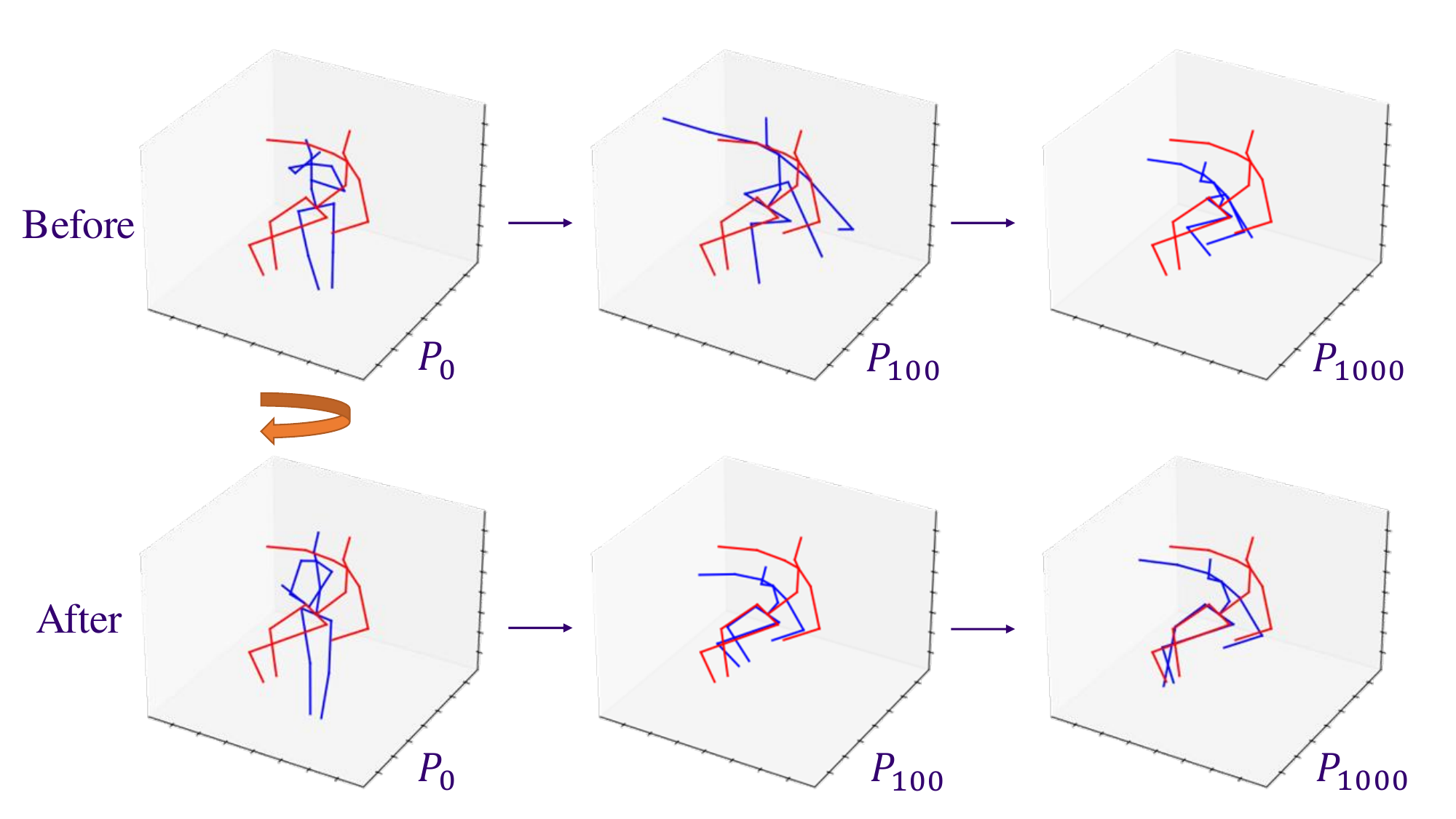}
    \caption{The rotation optimization in the initial pose optimizer finds the optimal initial pose and prevents the collapse of the estimated pose.}
    \label{fig:rotopt}
\end{figure}

\subsection{Initial Pose Optimizer}

Similar to other optimization-based 3D HPE methods\cite{bogo2016smplify, thomas2018joint, renshu2022unsupervised}, our optimization pipeline starts from an initial pose, $P_{init}$. As depicted in Fig~\ref{fig:rotopt}, the optimized pose may not suffice if the initial pose's orientation is significantly different from that of the target pose or even perpendicular to that of the target pose. Therefore, the initial pose optimizer is designed to find the optimal rotation matrix, $R_{0} \in SO(3)$, and translation, $T_{0}$, of $P_{init}$ by minimizing the re-projection error with 2D keypoints $p_{2d}$.

\begin{align}
    \argmin_{R_{0},T_{0}} \quad & \Big\|K(R_{0}P_{init} + T_{0}) - p_{2d}\Big\|_2 \\
    \textrm{s.t.} \quad & T_{min} \leq T_{0} \leq T_{max},
\end{align}

\noindent where $K$ is the camera intrinsic matrix. After the rotation, the $P_0 = R_oP_{init}$ is the optimal pose sent to the iterative optimization pipeline. As shown in Fig~\ref{fig:rotopt}, the rotation optimization aligns the initial poses with target 2D and 3D poses.

\subsection{Optimizer in the Loop}

In previous works\cite{bogo2016smplify, muller2021SMPLify-XMC, gu2020towards, thomas2018joint}, to optimize an accurate 3D human pose, it requires a lot of kinematic constraints, which call for a strong domain knowledge about human motion. Different from those previous works, as shown in Alg~\ref{alg:pipeline}, our iterative optimizer utilizes the denoising capability and learned human pose prior of the pose generation diffusion model to estimate an accurate 3D human pose with a simple yet effective optimization pipeline without any explicit kinematic constraint.

With a camera intrinsic matrix, $K$, 2D keypoints, $p_{2d}$, can be converted to 3D rays, $r \in R^{J\times3}$, based on perspective projection,

\begin{algorithm}[t]
\caption{ZeDO pipeline}\label{alg:pipeline}
\begin{algorithmic}
\Require $\text{Initial 3D pose}~P_{init}, \text{Target 2D pose}~p_{2d},$ \\ $\text{2D pose confidence scores}~C_{2d}, \text{Camera intrinsic}~K,$ \\ $\text{Diffusion timestamp}~t, \text{Pre-trained diffusion model}~\theta_{g}(P, t)$
\State $R_{0}, T_{0} \gets \argmin_{R_{0},T_{0}} \|K(R_{0}P_{init} + T_{0}) - p_{2d}\|_2$
\State \textcolor{gray}{// \textit{Initial Pose Optimization}}
\State $P_{0} \gets R_{0}P_{init}$
\State \textcolor{gray}{// \textit{Iterative Optimization and Denoising}}
\State $r \gets K^{-1}p_{2d}$
\State $\hat{r} \gets \frac{r}{\|r\|_2}$
\For{$i \gets 0$ to $n-1$}
    \If{$i < warmup$}
        \State $T_i \gets T_{0}$
    \Else
        \State $T_{i} \gets \argmin_{T_i} \|C_{2d}(K(P_{i} + T_{i}) - p_{2d})\|_2$
    \EndIf
    \State \textcolor{gray}{// \textit{Project 3D keypoints to rays}}
    \State $\widetilde{P}_{i} \gets \Big((P_{i} + T_{i}) \cdot \hat{r}\Big)\hat{r} - T_i$
    \State $P_{i+1} \gets \theta_g(\widetilde{P}_{i}, t(i))$
\EndFor
\State return $P_n$
\end{algorithmic}
\end{algorithm}

\begin{equation}
    r = K^{-1}p_{2d}, \hat{r} = \frac{r}{\|r\|_2}.
\end{equation}

\begin{figure}[t]
    \centering
    \includegraphics[width=0.8\linewidth]{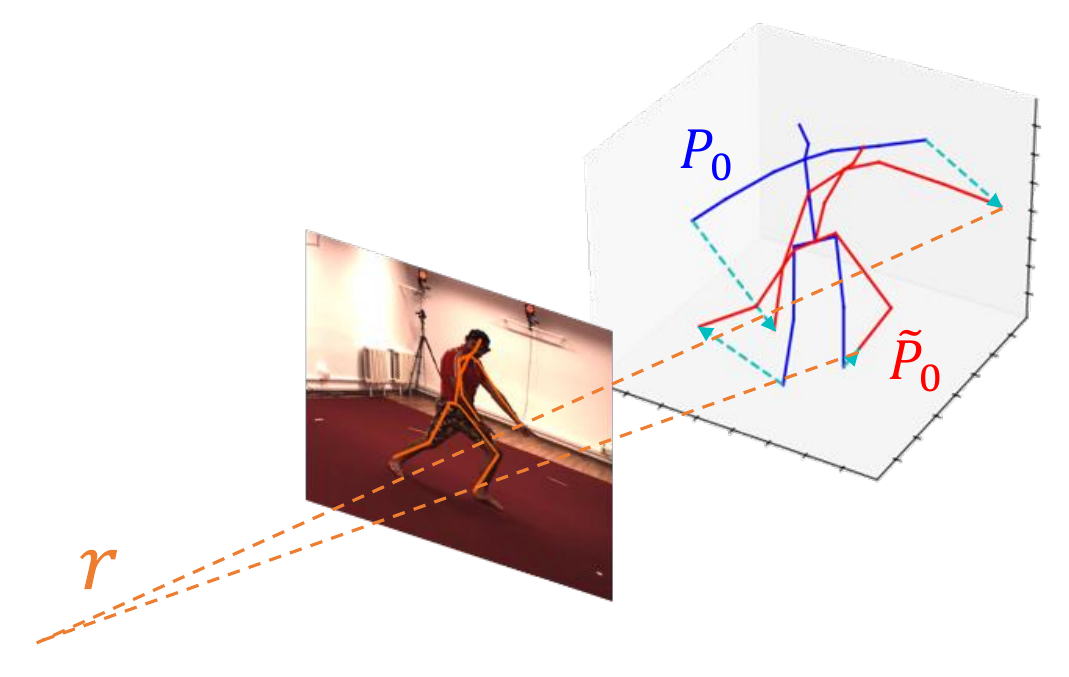}
    \caption{By projecting the $P_i$ to $r$, we minimize the 2D re-projection error and find an optimized pose $\widetilde{P}_i$.}
    \label{fig:rayproject}
\end{figure}

Intuitively, as shown in Fig~\ref{fig:rayproject}\ref{fig:rayproject}, projecting the 3D keypoints from $P_0$ to $r$ will minimize the 2D re-projection error and provide the estimated 3D human poses,

\begin{equation}
    \widetilde{P}_{0} \gets \Big((P_0 + T) \cdot \hat{r}\Big)\hat{r} - T.
\end{equation}

However, there are two problems: 1) Simply projecting 3D keypoints from $P_0$ to $r$ generates a noisy 3D human pose, which may not satisfy the kinematic constraints of the human body. 2) With different translations between $P_0$ and the camera, there are different projection results. 

In order to solve these two problems, we need to ensure the estimated 3D poses, $P_i$, are valid poses and inherently follow the kinematic constraints to find the optimal translation, $T_i$. This calls for our use of pose prior.

\noindent \textbf{Pose prior} Although there is no re-projection error from the optimized poses, the optimized poses may not satisfy the kinematic constraints. Therefore, in previous works, kinematic constraints are added to the pose optimization pipeline, and a complex joint optimization problem is designed to find optimal poses. However, in our pipeline, we take advantage of a pre-trained diffusion-based pose generation model to 'denoise' our optimized poses. As mentioned in DDPM\cite{2020DDPM}, DDIM\cite{song2020denoising} and Score Matching network\cite{song2020score}, the diffusion model is trained by maximizing likelihood, which aims at finding the most possible valid pose based on the input noisy pose. As a result, we use the diffusion-based pose generation model to find the optimal $P_{i+1}$ based on optimized pose $\widetilde{P}_i$, 

\begin{equation}
    P_{i+1} = \theta_g(\widetilde{P}_{i}, t(i)),
\end{equation}

which is different from other diffusion-based methods, like GFPose\cite{ci2023gfpose} and DiffPose\cite{gong2023diffpose}, $P = \theta(x, c, t)$, where $x$ is random noise, $c$ is pose condition and $t$ is timestamp.

However, during training, the reverse diffusion starts from the standard Gaussian noise, $\mathcal{N}\left(0, 1\right)$, but in our case, the generation model is utilized to denoise an optimized pose, which does not follow the standard Gaussian noise. Inspired by \cite{meng2021sdedit, zheng2022truncated}, we adopt truncated diffusion model inference, whose timestamp is truncated from the training timestamp during inference. In our case, the timestamp, $t$, is truncated as $t \in \left(0, 0.1\right]$, instead of $\left(0, 1\right]$.

\vspace{-0.5cm}

\paragraph{Find optimal translations} Optimal translations are derived from $T_o$ by minimizing the 2D re-projection error of $P_{i}$, depending on the current iteration number, since the optimized pose, $P_i$, is not reliable enough in the early iterations.

After certain iterations, the optimal translation is derived from the following:
\vspace{-0.45cm}

\begin{equation}
    T_{i} \gets \argmin_{T_i} \Big\|C_{2d}\Big(K(P_{i} + T_{i}) - p_{2d}\Big)\Big\|_2,
\end{equation}

\noindent where $C_{2d} \in R^{J}$ are the confidence scores of 2D keypoints, $p_{2d}$ and $K$ are the camera intrinsic matrices. Inspired by \cite{iskakov2019learnable}, $C_{2d}$ are helpful to guide the translation optimization. There is a closed-form solution to the optimization problem. The details can be found in the supplementary material.

By incorporating optimal translations and kinematic constraints, we can optimize the 3D human poses by iteratively projecting $P_i$ to $r$ and denoising the projected $P_{i+1}$ with the help of the diffusion model.

\section{Experimental Results}
\label{sec:exp}

In this section, we will introduce the experimental results of ZeDO on 3DPW\cite{3DPW}, Human3.6M\cite{H36M} and MPI-INF-3DHP\cite{3DHP} datasets. More results on Ski-Pose\cite{rhodin2018ski} datasets are included in the supplementary materials. Since ZeDO requires an initial pose as a starting point for optimization, and different initial poses may lead to different pose estimation results, we report our results in Mean Per Joint Position Error (MPJPE) with a single initial pose (single-hypothesis) or minimum MPJPE (minMPJPE) with multiple initial poses, in order to have fair comparison with previous multi-hypothesis 3D HPE methods\cite{ci2023gfpose, wehrbein2021probabilistic, sharma2019monocularmulti}.

\begin{table}[t]
\centering
\resizebox{0.9\linewidth}{!}{%
\setlength{\tabcolsep}{0.08cm}
\begin{tabular}{l | c | c | c | c }
\toprule
 Methods & CE & Opt & PA-MPJPE $\downarrow$ & MPJPE $\downarrow$\\
\midrule
  Kolotouros \etal\cite{SPIN} & & & 59.2 & 96.9 \\
  Kocabas \etal\cite{kocabas2020vibe} & & & 51.9 & 82.9 \\
  Kocabas \etal\cite{kocabas2021pare} & & & 46.4 & 74.7  \\
  Li \etal\cite{li2021hybrik} & & &  \underline{45.0} & \underline{74.1} \\
  Ma \etal\cite{ma2023virtualmarker} & & & \textbf{41.3} & \textbf{67.5} \\
\midrule
  Li \etal\cite{li2021hybrik} & \checkmark & & 50.9 & 82.0 \\
  Kocabas \etal\cite{kocabas2020vibe} & \checkmark & & 56.5 & 93.5 \\
  Kocabas \etal\cite{kocabas2021pare} & \checkmark & & 50.9 & 82.0  \\
  Gong \etal\cite{poseaug} & \checkmark & & 58.5 & 94.1 \\
  Gholami \etal\cite{gholami2022adaptpose} & \checkmark & & 46.5 & \underline{81.2} \\
  Chai \etal\cite{chai2023global} & \checkmark & & 55.3 & 87.7 \\
  Song \etal \cite{song2020learnedgradient} & \checkmark & & 55.9 & - \\
  Choutas \etal \cite{choutas2022FitMorphable} & \checkmark & & 52.2 & - \\
   \rowcolor[gray]{0.9}
   \textbf{ZeDO}~$(S=1, J=17)$ & \checkmark & \checkmark & \textbf{40.3} & \textbf{69.7}\\
   \rowcolor[gray]{0.9}
   \textbf{ZeDO}~$(S=1, J=14)$ & \checkmark & \checkmark & \underline{43.1} & 76.6\\
\bottomrule
\end{tabular}
}
\caption{Cross-domain evaluation results on 3DPW dataset. CE stands for cross-domain evaluation, and Opt means optimization-based method. Ground truth 2D poses are used.}
\label{tab:3DPW}
\end{table}

\subsection{Datasets} 

\noindent \textbf{Human3.6M}\cite{H36M} is the most widely used single-person 3D pose benchmark with more than 3.6 million frames and corresponding 3D human poses. The dataset is collected within a $4\text{ m} \times 3\text{ m}$ indoor environment, with 11 professional actors (6 males and 5 females) performing 17 distinct actions such as discussion, smoking, capturing photographs, posing, greeting, and talking on the phone. Following the convention of previous works\cite{videopose3d, SemGCN, evoskeleton} for fair comparison, we use the S1, 5, 6, 7, and 8 as the training dataset and evaluate the model on S9 and S11. 

\noindent \textbf{MPI-INF-3DHP}\cite{3DHP} is a large 3D human pose dataset with more than 1.3 million frames captured indoors and outdoors. The 3DHP dataset captures poses of 8 actors, consisting of 4 males and 4 females with 8 different actions each, encompassing a range of activities from simple actions like walking and sitting to more complex exercise poses and dynamic movements. Following \cite{poseaug}, we use a sampled $2929$ frame test dataset.

\noindent \textbf{3DPW}\cite{3DPW} is the first dataset in-the-wild with accurate 3D poses for evaluation. Compared with Human3.6M and MPI-INF-3DHP, 3DPW focuses on outdoor scenarios and captures videos with static and moving cameras. There are 60 video sequences captured in the dataset with 18 different actors. Following \cite{chai2023global}, we test ZeDO on 3DPW only for cross-dataset evaluation.

\begin{table}[t]
\centering
\setlength{\tabcolsep}{0.08cm}
\begin{tabular}{ l | c | c }
\toprule
 \textbf{Learning} Methods & MPJPE $\downarrow$ & PA-MPJPE $\downarrow$\\
\midrule
  Martinez \etal\cite{simplebaseline} & 62.9 & 47.7\\
  Zhao \etal\cite{SemGCN} & 57.6 & - \\
  Pavllo \etal\cite{videopose3d} $(f=1)$ & 52.7 & 40.9 \\
  Li \etal\cite{CLIFF} &\underline{47.1} & 32.7 \\
  Gong \etal\cite{poseaug} & 50.2 & 39.1\\
  Gong \etal\cite{gong2023diffpose} $(f=1)$ & 49.7 & \underline{31.6}\\
  Ci \etal\cite{ci2023gfpose}~$(S=1)$ & 51.0 & - \\
  Ci \etal\cite{ci2023gfpose}~$(S=10)$ & \textbf{45.1} & \textbf{30.5} \\
\midrule
\midrule
 \textbf{Optimization} Methods & MPJPE $\downarrow$ & PA-MPJPE $\downarrow$\\
\midrule
   Wang \etal\cite{wang2014robust} & 88.0 & - \\
   Bogo \etal\cite{bogo2016smplify} & 82.3 & - \\
   Li \etal\cite{li2015maximum} & 78.6 & - \\
   Gu \etal\cite{gu2020towards} & 77.2 & - \\
   Song \etal\cite{song2020learnedgradient} & - & 56.4 \\
  \rowcolor[gray]{0.9}
   \textbf{ZeDO}~$(S=1)$ & 65.7 & 49.0\\
   \rowcolor[gray]{0.9}
   \textbf{ZeDO}~$(S=10)$ & \underline{57.3} & \underline{45.1}\\
   \rowcolor[gray]{0.9}
 \textbf{ZeDO}~$(S=50)$ & \textbf{51.4} & \textbf{42.1}\\
\bottomrule
\end{tabular}
\caption{3D HPE quantitative results on Human3.6M dataset. $S$ indicates the number of hypotheses. All results are reported in millimeters (mm). The pose generation model is trained on Human3.6M. Detected 2D poses by Stacked Hourglass are used.}
\label{tab:h36msame}
\end{table}

\subsection{Training and Inference Details}

We pre-train our 3D pose generation model for 5000 epochs on one NVIDIA A100 with a batch size of 50k, a learning rate of $2e^{-4}$ with an Adam optimizer. The training schedule comes with warmup in the first 5k iterations and cosine learning rate decay in the following iterations. As \cite{song2020score}, the timestamp, $t$, during the forward or reverse diffusion process, is uniformly sampled from $\left(0,1\right]$. All 3D human poses are normalized to pelvis-related coordinates during training and inference. To improve the robustness of the model, we apply flip and rotation data augmentation during training. For cross-domain evaluation, we pre-train the pose generation model in a different dataset and directly test the optimization pipeline without any fine-tuning.

During inference, the pipeline supports single or multiple initial poses. For the initial pose optimizer, we limit the rotation axis to the z-axis only for better performance. We set the number of warmup iterations as $200$, and the number of total iterations as $1000$. The timestamp, $t$, is uniformly sampled from $\left(0, 0.1\right]$. The initial poses are sampled from training sets of Human3.6M\cite{H36M} or 3DHP\cite{3DHP} by the K-Means algorithm. For different numbers of hypotheses, we run K-Means with different numbers of clusters.

\begin{table}[t]
\centering
\resizebox{\linewidth}{!}{%
\setlength{\tabcolsep}{0.08cm}
\begin{tabular}{l | c | c | c | c | c}
\toprule
 Methods & CE & Opt & MPJPE $\downarrow$ & PCK $\uparrow$ & AUC $\uparrow$\\
\midrule
  Mehta \etal\cite{mehta2017vnect} & & & 124.7 & 76.6 & 40.4\\
  Martinez \etal\cite{simplebaseline} & & & 84.3 & 85.0 & 52.0\\
  Pavllo \etal\cite{videopose3d} $(f=1)$ & & & 86.6 & - & - \\
  Li \etal\cite{li2022mhformer} $(f=9)$ & & & 58.0 & 93.8 & 63.3 \\
  Zhang \etal\cite{zhang2022mixste} $(f=1)$ & & & \underline{57.9} & 94.2 & 63.8 \\
  \rowcolor[gray]{0.9}
   \textbf{ZeDO}~$(S=1)$ & & \checkmark & 86.5 & 82.6 & 53.8\\
   \rowcolor[gray]{0.9}
   \textbf{ZeDO}~$(S=50)$ & & \checkmark & \textbf{55.2} & 93.0 & 65.6\\
\midrule
Kanazawa \etal\cite{kanazawa2018HMR} & \checkmark & & 113.2 & 77.1 & 40.7\\
  Ci \etal\cite{ci2023gfpose} & \checkmark & & - & 86.9 & - \\
  Gong \etal\cite{poseaug} & \checkmark & & 73.0 & 88.6 & 57.3\\
  Gholami \etal\cite{gholami2022adaptpose} & \checkmark & & 68.3 & 90.2 & 59.0\\
  Chai \etal\cite{chai2023global} & \checkmark & & \textbf{61.3} & 92.1 & 62.5 \\
  M{\"u}ller \etal\cite{muller2021SMPLify-XMC} & \checkmark & \checkmark & 101.2 & - & - \\
   \rowcolor[gray]{0.9}
   \textbf{ZeDO}~$(S=1)$ & \checkmark & \checkmark & 99.9 & 81.8 & 50.9\\
   \rowcolor[gray]{0.9}
   \textbf{ZeDO}~$(S=50)$ & \checkmark & \checkmark &\underline{69.9} & 90.2 & 58.8 \\
\bottomrule
\end{tabular}
}
\caption{3D HPE quantitative results on 3DHP dataset. CE stands for cross-domain evaluation, and Opt means optimization-based method. Ground truth 2D poses are used.}
\label{tab:3DHP}
\end{table}

\subsection{Results}

\noindent \textbf{Results on 3DPW}. 3DPW is a challenging in-the-wild dataset, compared with Human3.6M and MPI-INF-3DHP datasets. 3DPW focuses on outdoor scenarios with both static and moving cameras. For cross-domain evaluation on the 3DPW dataset, we pre-train the pose generation model on the Human3.6M dataset and inference on the 3DPW dataset without any fine-tuning. On 3DPW, we find some of the previous works\cite{li2021hybrik, kocabas2021pare} evaluate on 14 Leeds Sports Pose (LSP)\cite{johnson2010lsp} keypoints, while others \cite{poseaug, gholami2022adaptpose, chai2023global} evaluate on 17 Human3.6M keypoints, and some other works\cite{SPIN} do not explain clearly which keypoints they use. In Table~\ref{tab:3DPW}, we report the results of both 14 and 17 keypoints for a fair comparison. We achieve SOTA performance,  PA-MPJPE $40.3$mm. with a single hypothesis. 

\noindent \textbf{Results on Human3.6M}. Although ZeDO is a Zero-shot Diffusion-based Optimization pipeline, ZeDO achieves comparable results with learning-based. Following \cite{ci2023gfpose}, we report the minMPJPE of multi-hypothesis, i.e., $S$ number of initial poses, in inference and use the detected 2D poses as input. As shown in Table~\ref{tab:h36msame}, on the Human3.6M dataset, we obtain $51.4$mm in minMPJPE with $S=50$, which is comparable with SOTA learning-based methods, while ZeDO does not train with any 2D-3D or image-3D pairs. Compared with other optimization-based methods, ZeDO outperforms previous works by a large margin, even with $S=1$. In this experiment, we train the pose generation model on the training set of Human3.6M.

\noindent \textbf{Results on 3DHP}. Following previous works\cite{poseaug, gholami2022adaptpose, chai2023global}, we use ground truth 2D poses as input. As shown in Table~\ref{tab:3DHP}, with $S=50$, ZeDO even outperforms the learning-based methods by $2.7$mm in minMPJPE. In the cross-domain evaluation, we achieve SOTA performance as $67.9$mm in minMPJPE and outperform the optimization-based method by a large margin, with $S=50$.

\vspace{-0.5em}

\section{Discussion}
\label{sec:discuss}

\begin{table*}[t!]
\centering
\resizebox{0.8\linewidth}{!}{%
\begin{tabular}{ c | c | c  c  c c | c  c | c  c }
\toprule
  \multirow{2}{*}{Dataset} & \multirow{2}{*}{Diff Model} & \multirow{2}{*}{RO} & \multirow{2}{*}{WU} & \multirow{2}{*}{RA} & \multirow{2}{*}{GT} & \multicolumn{2}{c|}{$S=1$} & \multicolumn{2}{c}{$S=50$} \\
  & & & & & & MPJPE $\downarrow$ & PA-MPJPE $\downarrow$ & MPJPE $\downarrow$ & PA-MPJPE $\downarrow$\\
\midrule
  H36M & H36M & & & & & 75.0 & 52.7 & 53.4 & 42.7\\
  H36M & H36M & \checkmark & & & & 77.2	&53.7	&	52.7	&42.4\\
  H36M & H36M & \checkmark & \checkmark & & & \textbf{65.7}~\footnotesize{(\textcolor{darkgreen}{$9.3\downarrow$})}	&\textbf{49.0}~\footnotesize{(\textcolor{darkgreen}{$3.7\downarrow$})}	&	\textbf{51.4}~\footnotesize{(\textcolor{darkgreen}{$2.0\downarrow$})}&	\textbf{42.1}~\footnotesize{(\textcolor{darkgreen}{$0.6\downarrow$})}\\
  H36M & H36M & \checkmark & \checkmark & \checkmark & & 69.5 & 51.4 & 52.9 & 42.5\\
  H36M & H36M & \checkmark & \checkmark & & \checkmark & 50.1&	35.8	&	37.0&	27.5\\
\midrule
  3DHP & H36M &  &  &  & \checkmark & 148.3 &	88.8 & 93.4 & 59.0 \\
  3DHP & H36M & \checkmark & \checkmark &  & \checkmark & 113.8 &	74.1 & 80.1 &	56.0 \\
  3DHP & H36M & \checkmark  & \checkmark  & \checkmark  & \checkmark & \textbf{99.9}~\footnotesize{(\textcolor{darkgreen}{$48.4\downarrow$})}	& \textbf{67.9}~\footnotesize{(\textcolor{darkgreen}{$20.9\downarrow$})}	&	\textbf{69.9}~\footnotesize{(\textcolor{darkgreen}{$23.5\downarrow$})}    &	\textbf{49.0}~\footnotesize{(\textcolor{darkgreen}{$10.0\downarrow$})} \\
\midrule
  3DHP & 3DHP & \checkmark & \checkmark &  & \checkmark & 86.5&	55.9 & 55.2 & 38.6 \\
\bottomrule
\end{tabular}
}
\caption{The ablation study results of ZeDO. RO stands for rotation optimization as the initial pose optimization. WU denotes the warmup iterations. RA is the rotation data augmentation for training the pose generation model. The dataset name under the Dataset column is the testing dataset, and the name under the Diff Model column is the dataset used for diffusion model pre-training.}
\label{tab:ablation}
\end{table*}

\subsection{Ablation Studies}


\noindent \textbf{Different diffusion-based pose generation models.} As shown in table~\ref{tab:diffbb}, we try to evaluate the cross-domain performance of our pipeline with different diffusion-based backbones on the 3DPW. We test the Score Matching Network\cite{song2020score}, DDPM\cite{2020DDPM}, and DDIM\cite{song2020denoising} models trained on the Human3.6M dataset and keep all other settings the same. It turns out that DDIM also achieves comparable performance in terms of PA-MPJPE and even lower MPJPE compared with the Score Matching Network we report above. The outcome validates the generality and viability of our idea, regardless of the specific structure of the diffusion backbone.

\begin{table}[t]
\centering
\resizebox{0.75\linewidth}{!}{%
\setlength{\tabcolsep}{0.08cm}
\begin{tabular}{ l | c  c }
\toprule
 Diffusion Backbone  & PA-MPJPE $\downarrow$ & MPJPE $\downarrow$\\
\midrule
  Score Matching \cite{song2020score}   &  \textbf{40.3} & \underline{69.7} \\ 
  \midrule
  DDIM\cite{song2020denoising}   & \underline{40.4} & \textbf{67.9}\\
  DDPM\cite{nichol2021improved}   & 51.7  & 81.3\\ 
\bottomrule
\end{tabular}
}
\caption{Different diffusion backbone 3D HPE quantitative results on 3DPW dataset.}
\label{tab:diffbb}
\end{table}

\begin{table}[b]
\centering
\resizebox{0.8\linewidth}{!}{%
\begin{tabular}{ c | c | c c }
\toprule
 Dataset & Diff Model & MPJPE $\downarrow$ & PA-MPJPE  $\downarrow$\\
 \midrule
 H36M & H36M & 37.0 & 27.5 \\
 H36M & mixed & \textbf{35.7} & \textbf{26.5} \\
 \midrule
 3DHP & H36M & 69.9 & 49.0 \\
 3DHP & 3DHP & 55.2 & 38.6 \\
 3DHP & mixed & \textbf{52.4} & \textbf{37.7} \\
\bottomrule
\end{tabular}
}
\caption{The pose generation models trained on mixed datasets (Human3.6M + 3DHP) achieves the best performance with $S=50$, as well as better generalization. GT 2D keypoints are used.}
\label{tab:mixed}
\end{table}

\noindent \textbf{How does initial pose optimizer help ZeDO?} The initial pose optimizer is designed to align the initial pose with the target 2D pose by rotation for better initialization. As shown in Table~\ref{tab:ablation}, the combination of rotation optimization and warmup iterations reduces the MPJPE by $9.3$ mm when $S=1$ and the minMPJPE by $2.0$mm when $S=50$, on the Human3.6M dataset. When $S=50$, hypotheses cover lots of different pose orientations,  resulting in  the relatively smaller improved performance from the initial pose optimizer when $S$ is larger. On the 3DHP dataset, the initial optimization further improves the performance by $34.5$ mm in MPJPE when $S=1$ since 3DHP contains more complex 3D human poses in different orientations than Human3.6M. The initial pose optimization is able to generate a reliable optimized initial pose and an optimal translation as the warmup translation for the following iterative optimization pipeline.

\noindent \textbf{Does data augmentation help the performance?} In ZeDO, the diffusion model is pre-trained for pose generation. However, the 3D pose distributions vary across different datasets. To ensure the pre-trained diffusion model can be adapted to different datasets, we utilize rotation and flip data augmentation during training. As expected, the data augmentation significantly improves the performance in cross-domain evaluation by $13.9$ mm in MPJPE, shown in Table~\ref{tab:ablation}. 


\noindent \textbf{Boost the performance further by mixing dataset.} According to Table~\ref{tab:mixed}, the pose generation model trained on the mixed datasets (Human3.6M + 3DHP) improves the performance of ZeDO  by $1.3$mm on Human3.6M and $2.8$mm on 3DHP while using the same estimation algorithm.

\begin{figure}[t]
    \centering
    \includegraphics[width=0.9\linewidth]{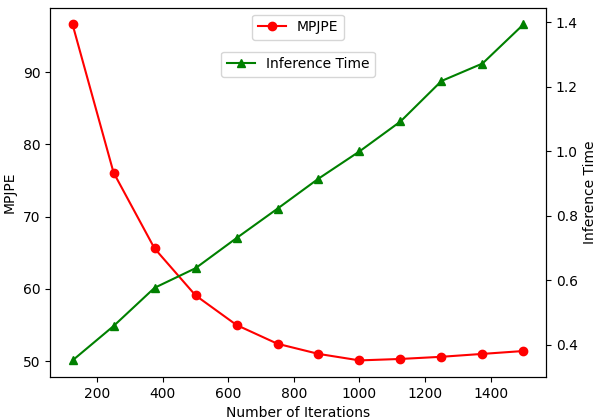}
    \caption{The MPJPE and inference time with different numbers of iterations on the Human3.6M dataset with ground truth 2D keypoints.}
    \label{fig:iterations}
\end{figure}


\noindent \textbf{How to pick the initial pose?} We utilize K-Means in our experiments to select anchor poses from the training set as initial poses. K-Means effectively finds the most representative poses in the training set, making it superior to other sampling strategies, as shown in Table~\ref{tab:sample}. Random Sampling randomly samples a pose from the training set, whereas Random Generation is generated by the pre-trained pose generation model.

\begin{table}[b]
\centering
\begin{tabular}{ c | c | c c }
\toprule
 Sampling & MPJPE $\downarrow$ & PA-MPJPE  $\downarrow$\\
 \midrule
 Random Sampling & 78.2 & 51.2 \\
 Random Generation & 70.4 & 46.0 \\
 K-Means & \textbf{50.1} & \textbf{35.8} \\
\bottomrule
\end{tabular}
\caption{Results on Human3.6M dataset with different sampling strategies when $S=1$. Ground truth 2D keypoints are used.}
\label{tab:sample}
\end{table}


\noindent \textbf{What is the best number of optimization iterations?} In ZeDO, the number of diffusion optimization iterations is set to $1000$. Intuitively, increasing the number of iterations can enhance performance but may suffer the inference speed. In Fig~\ref{fig:iterations}, as expected, the inference time increases linearly with respect to the number of iterations. However, the figure shows that the best performance is achieved when the number of iterations is around $1000$. With the number of iterations exceeding $1000$, there is no performance gain, and the inference speed decreases.

\subsection{Limitations} 
\begin{figure}[t]
    \centering
    \includegraphics[width=0.9\linewidth]{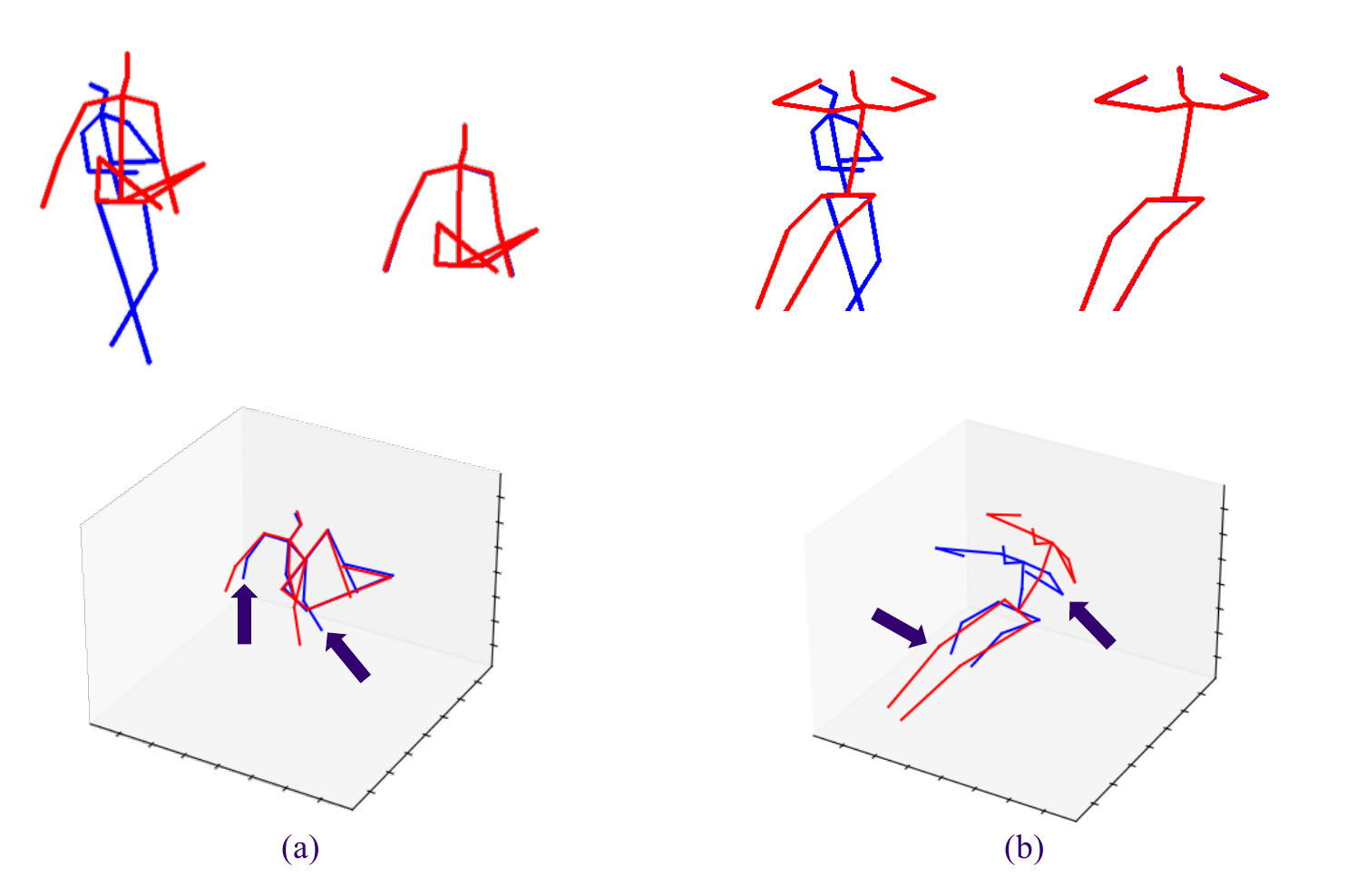}
    \caption{The failure cases of our method, because of the one-to-many issue in 3D HPE.}
    \label{fig:failure}
\end{figure}

Although ZeDO achieves state-of-the-art performance in various benchmarks and settings, several limitations still need to be further explored. 1) Similar to other optimization-based approaches, the optimizer in the loop requires camera intrinsic parameters. 2) Since our method is based on minimizing the 2D re-projection error, we are not able to solve the ambiguity of the depth and scale without additional information like bone length or height. 3) For an identical 2D human pose, there are multiple 3D human poses matched. Without image or temporal information, the 1-to-many mapping issue cannot be resolved by single frame lifting methods, as shown in Fig~\ref{fig:failure}.

\section{Conclusion}
\label{sec:conclusion}

In this paper, we propose ZeDO, a Zero-shot Diffusion-based Optimization pipeline for 3D HPE. To the best of our knowledge, we are the first to introduce the diffusion model to the optimization-based method in the 3D HPE task. We leverage the pre-trained diffusion-based 3D human pose generation model and can optimize target 3D poses in the loop. To be specific, an optimizer that calculates the optimal translation is used iteratively with denoising steps in the diffusion model. Compared to other prior arts, ZeDO achieves state-of-the-art performance on Human3.6M, MPI-INF-3DHP, and 3DPW datasets, even with cross-dataset evaluation. In the future, we plan to further improve ZeDO by modifying the diffusion model and solving the limitations listed in Sec~\ref{sec:discuss}. It is our wish that this optimization method could become a common paradigm beyond end-to-end lifting networks in 3D human pose estimation tasks.

\newpage
{\small
\bibliographystyle{ieee_fullname}
\bibliography{egbib}
}

\newpage
\input{new_supp}

\end{document}

%% file: new_supp.tex
{\centering\section*{Supplementary Material}}

\begin{figure}[h]
    \centering
    \includegraphics[width=\linewidth]{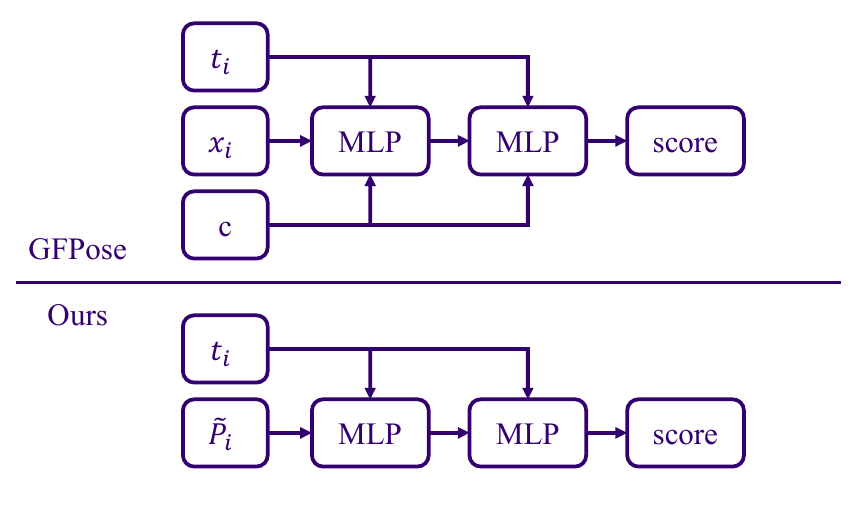}
    \caption{The architecture of GFPose and our diffusion model. Compared with GFPose, there is no pose condition $c$ as input, and the noise $x_i$ is replaced by optimized pose $\widetilde{P}_i$.}
    \label{fig:comparison}
\end{figure}

\section*{A. Architecture Difference with GFPose}

As shown in Fig~\ref{fig:comparison}, compared with GFPose, there is no pose condition $c$ as input, and the noise $x_i$ is replaced by optimized pose $\widetilde{P}_i$. Our model is not the same as the GFPose. We utilize Score Matching Network to build our human pose generation model.

\section*{B. Initial Pose Optimizer}

In the initial pose optimizer, our optimization target is

\begin{align}
    \argmin_{R_{0},T_{0}} \quad & \Big\|K(R_{0}P_{init} + T_{0}) - p_{2d}\Big\|_2 \\
    \textrm{s.t.} \quad & T_{min} \leq T_{0} \leq T_{max}.
\end{align}

To solve this optimization problem, we use the Adam optimizer, with the learning rate as $0.1$ and optimization iterations as $500$. Instead of optimizing the $3\times3$ rotation matrix, we optimize $R_{0}$ based on quaternion to ensure the generated $R_{0} \in \text{SO}(3)$.

\section*{C. Optimize Translation}

As described in Sec 3, there is a closed-form solution of translation optimization. The optimization target is

\begin{equation}
    \argmin_{T_i} \quad \Big\|C_{2d}\Big(K(P_{i} + T_{i}) - p_{2d}\Big)\Big\|_2.
\end{equation}

The target can be solved by formalizing to

\begin{align*}
    \argmin_{T_i} & \quad \Big\|C_{2d}\Big(K(P_{i} + T_{i}) - p_{2d}\Big)\Big\|_2 \\
    = \argmin_{T_i} & \quad \Big\|AT_i - b\Big\|_2 \\
    \text{where,} \\
    A &= \begin{bmatrix}
        -C_{2d, 0} & 0 & C_{2d, 0}r_{(0, 0)}\\
        0 & -C_{2d, 0} & C_{2d, 0}r_{(0, 1)} \\
        & \vdots & \\
        -C_{2d, J} & 0 & C_{2d, J}r_{(J, 0)}\\
        0 & -C_{2d, J} & C_{2d, J}r_{(J, 1)} \\
        \end{bmatrix}\\
    b &= \begin{bmatrix}
        C_{2d, 0}(P_{i(0, 0)} - P_{i(0, 2)}r_{(0, 0)})\\
        C_{2d, 0}(P_{i(0, 1)} - P_{i(0, 2)}r_{(0, 1)})\\
        \vdots \\
        C_{2d, J}(P_{i(J, 0)} - P_{i(J, 2)}r_{(J, 0)})\\
        C_{2d, J}(P_{i(J, 1)} - P_{i(J, 2)}r_{(J, 1)})
        \end{bmatrix} \\
    r &= \frac{K^{-1}p_{2d}}{\|K^{-1}p_{2d}\|} 
\end{align*}

The optimization target can be solved as

\begin{equation}
    T_i = (A^TA)^{-1}A^Tb
\end{equation}

\begin{table}[t]
\centering
\resizebox{\linewidth}{!}{%
\setlength{\tabcolsep}{0.08cm}
\begin{tabular}{ l |l| c  c }
\toprule
 Dataset & Methods & MPJPE $\downarrow$ & PA-MPJPE $\downarrow$\\
\midrule
\multirow{2.4}*{3DPW\cite{3DPW}} & VPose3D($f$=1)\cite{videopose3d}& 75.9 & 48.8 \\ 
   \cmidrule{2-4}
    & + ZeDO & \textbf{70.2}\textcolor{green}{(-5.7)} & \textbf{39.3}\textcolor{green}{(-9.5)}\\
  \midrule
   \multirow{2.4}*{H36M\cite{H36M}} & VPose3D($f$=1)& 39.2 & 30.4 \\ 
   \cmidrule{2-4}
    & + ZeDO & \textbf{38.7}\textcolor{green}{(-0.5)} & \textbf{27.8}\textcolor{green}{(-2.6)}\\
  \midrule

  \multirow{2.4}*{3DHP\cite{3DHP}} & VPose3D($f$=1)& 89.1  & 60.5 \\
  \cmidrule{2-4}
    & + ZeDO & \textbf{78.2}\textcolor{green}{(-10.9)} & \textbf{51.9}\textcolor{green}{(-8.6)}\\
\bottomrule
\end{tabular}
}
\caption{Refinement quantitative results on all three datasets. Our method could further reinforce the performance of the traditional 2D-3D lifting model VideoPose3D\cite{videopose3d}, in which $f=1$ represents the single frame scenario. All experiments are $S=1$. GT 2D poses are used.}
\label{tab:composeaug}
\end{table}

\section*{D. 3D Pose Refinement Results}
ZeDO not only has the capacity of denoising pre-defined pose priors but also refines outputs produced by existing 2D-3D lifting networks. In order to validate its effectiveness, we conduct comparative experiments pitting single frame VideoPose3D~\cite{videopose3d} against our model, aiming to prove that our model could further enhance performance. As demonstrated in ~\ref{tab:composeaug}, we run our mixed-dataset-trained model by taking the keypoint outputs from VideoPose3D as initialization. As a result, we attain lower MPJPE performance on all the datasets, which proves ZeDO's outstanding refinement ability.

\section*{E. Results on Ski-Pose Dataset}
Ski-Pose\cite{rhodin2018ski} is a dataset focusing on ski data, which provides labels for the skiers’ 3D poses in each frame and their projected 2D pose in all 20k images. We tested our model as the cross-dataset evaluation on Ski-Pose dataset. As shown in Table~\ref{tab:ski}, we achieve SOTA as PA-MPJPE $81.0$mm with the single hypothesis.

\begin{table}[h]
\centering
\setlength{\tabcolsep}{0.08cm}
\begin{tabular}{ l | c | c  c }
\toprule
 Methods & CE & PA-MPJPE $\downarrow$ & MPJPE $\downarrow$\\
\midrule
  Rhodin \etal\cite{rhodin2018ski} & & 85.0 & - \\ 
  Wandt \etal\cite{wandt2021canonpose} & & 89.6 & 128.1 \\ 
  \midrule
  Pavllo \etal\cite{videopose3d} & \checkmark & 88.1 & 106.0\\
  Gong \etal\cite{poseaug} & \checkmark & 83.5 & 105.4\\
  Gholami\cite{gholami2022adaptpose} & \checkmark & 83.0 & \underline{99.4} \\
  \rowcolor[gray]{0.9}
   \textbf{ZeDO}~$(S=1)$ & \checkmark & \underline{81.0} & 106.3\\
   \rowcolor[gray]{0.9}
 \textbf{ZeDO}~$(S=50)$ & \checkmark & \textbf{56.8} & \textbf{74.2}\\
\bottomrule
\end{tabular}
\caption{3D HPE quantitative results on Ski-Pose dataset. $S$ indicates the number of hypotheses. All results are reported in millimeters (mm). The pose generation model is trained on Human3.6M. GT 2D poses are used.}
\label{tab:ski}
\end{table}

\section*{F. In Comparison to Unsupervised Methods}
We also compared our results with unsupervised methods on the Human3.6m and 3DPW datasets, as shown in Table~\ref{tab:unsupervisedH36M} and ~\ref{tab:unsupervised3DPW}. Here, we only applied backbones trained on the Human3.6m dataset for evaluation. Apparently, our method outperforms all of the previous SOTA methods. 

\begin{table}[t]
\centering
\resizebox{\linewidth}{!}{%
\setlength{\tabcolsep}{0.08cm}
\begin{tabular}{ l |l| c  c }
\toprule
 Supervision & Methods & PA-MPJPE $\downarrow$ & N-MPJPE $\downarrow$\\
\midrule
 \multicolumn{2}{l}{\textbf{GT}}   \\
\midrule
Unsupervised & Chen\cite{ChenUnsupervised}& 58.0 & - \\ 
  
    & \cite{ChenUnsupervised}reimplemented by \cite{UnsupervisedRe} & 46.0 & - \\
  
    & Yu\cite{UnsupervisedRe}(temporal)& 42.0 & 85.3 \\ 

    & ElePose\cite{elepose}  &36.7 & 64.0\\
   \midrule
    & ZeDO $(S=1)$ & \textbf{35.8} & \textbf{46.9} \\

\midrule
\multicolumn{2}{l}{\textbf{DT}}  \\
\midrule

Unsupervised & Kundu\cite{kundu2020self}& 62.4 & - \\ 
  
    & Kundu\cite{Kundu2020KinematicStructurePreservedRF} & 63.8 & - \\
  
    & Chen\cite{UnsupervisedRe}& 68.0 & - \\ 

    & Yu\cite{UnsupervisedRe}  &52.3 & 92.4\\
    & ElePose\cite{elepose}  &50.2 & 74.4\\
   \midrule
    & ZeDO $(S=1)$ & \textbf{49.0} & \textbf{63.6} \\
\bottomrule
\end{tabular}
}
\caption{Quantitative results in comparison with unsupervised methods on Human3.6m dataset. The top table illustrates the results using GT 2D keypoints, and the bottom shows the results of detected 2D inputs. Our model attains top one performance among all unsupervised methods.}
\label{tab:unsupervisedH36M}
\end{table}

\begin{table}[t]
\centering
\resizebox{\linewidth}{!}{%
\setlength{\tabcolsep}{0.08cm}
\begin{tabular}{ l |l| c c c c }
\toprule
 Supervision & Methods & PA-MPJPE $\downarrow$ & N-MPJPE $\downarrow$ \\
\midrule

     Unsupervised & ElePose\cite{elepose}  &64.1 & 93.0  \\
   \midrule
    & ZeDO ($S=1$,$J=17$)& \textbf{40.3} &  \textbf{60.8}\\
  
\bottomrule
\end{tabular}
}
\caption{Quantitative results in comparison with unsupervised methods on Dataset 3DPW. GT 2D poses are used. The number of joints is 17.}
\label{tab:unsupervised3DPW}
\end{table}

\section*{G. Model Hyperparameter}

Crucial training and inference hyperparameters are displayed in Table \ref{tab:hyperparameter}. 

\begin{table}[t]
\centering
\setlength{\tabcolsep}{0.08cm}
\begin{tabular}{ l |   c }
\toprule
 Hyperparameter & \\
\midrule
    Batch Size & 1024\\
    
    Training Epoch &  2000 \\

    Training Optimizer &  Adam\cite{kingma2014adam} \\

    Training Learning rate & 2e-4\\

    Training Warmup Iterations & 5000 \\

    Training $\beta{_1}$ & 0.9 \\
    
   Training $\beta{_2}$ & 0.999 \\
   Inference timestamp $t$  & (0, 0.1]\\
  
   Inference Iteration Steps&  1000 \\  

   Inference Optimizer & Adam \\

   Optimization Ratotaion Axis & Z \\

   $T_{min}$ & 1.6m \\ 

   $T_{max}$ & 16m \\
\bottomrule
\end{tabular}
\caption{Important hyperparameters of training and inference on the 3DPW dataset. }
\label{tab:hyperparameter}
\end{table}

\begin{figure*}[t]
    \centering
    \includegraphics[width=\linewidth]{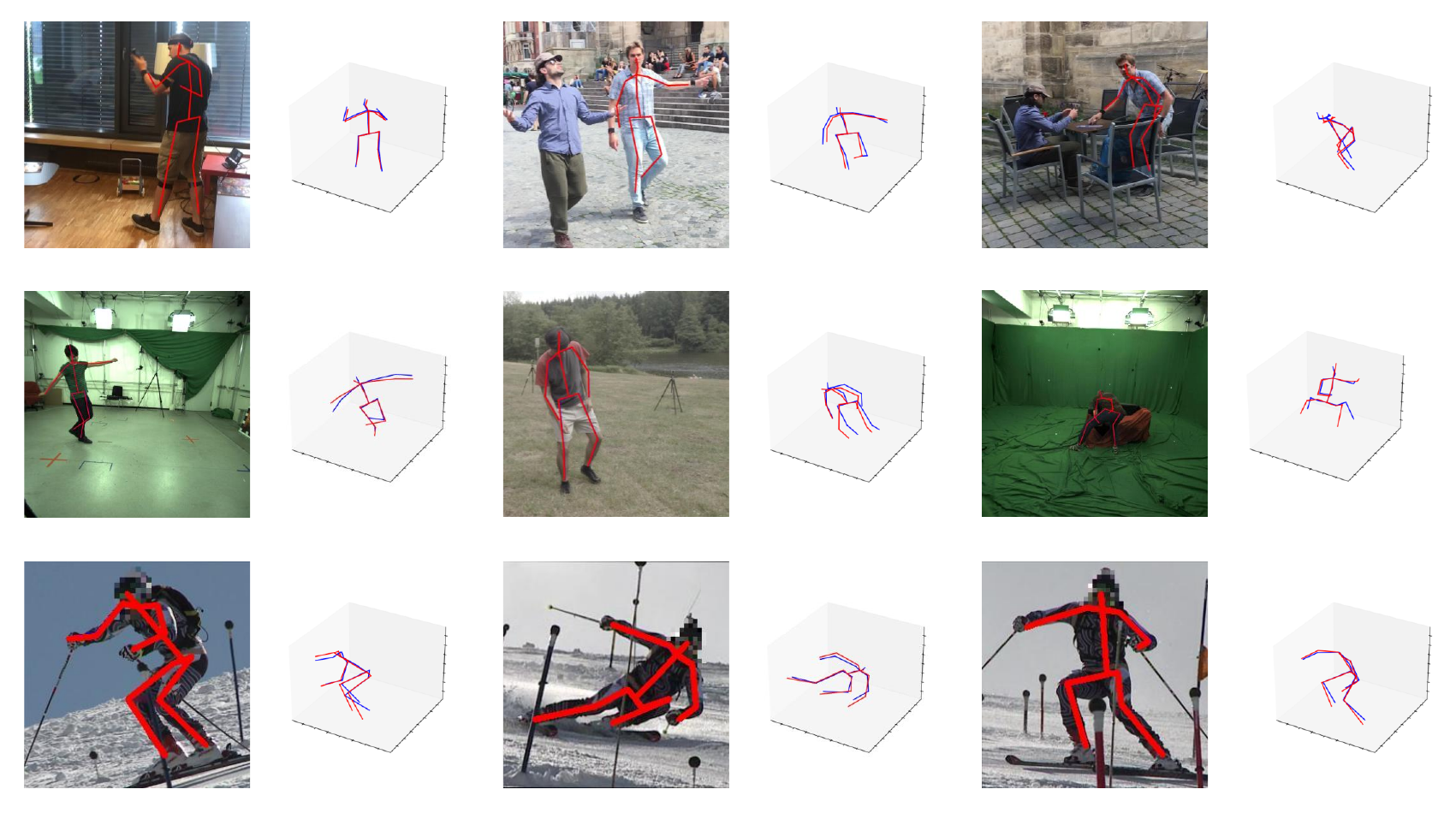}
    \caption{3D HPE qualitative results on 3DPW, MPI-INF-3DHP and Ski-Pose datasets. First row: 3DPW. Second row: 3DHP. Third row: Ski-Pose.}
    \label{fig:3DPW}
\end{figure*}